\documentclass[APA,Garamond1COL]{WileyNJDv5} 

\articletype{Article Type}%

\received{Date Month Year}
\revised{Date Month Year}
\accepted{Date Month Year}
\journal{Journal}
\volume{00}
\copyyear{2023}
\startpage{1}

\usepackage{hyperref}

\usepackage{amsfonts, amsmath, amssymb, amsthm, constants,comment}
\usepackage{mathrsfs}%
\usepackage{bbm,dsfont}

\DeclareMathOperator*{\rank}{rank}
\DeclareMathOperator*{\sign}{sign}

\DeclareMathOperator*{\argmax}{arg\, max}
\DeclareMathOperator*{\argmin}{arg\, min}

\def \P {\mathbb{P}}

\def \moo| {\langle}
\def \< {\langle }
\def \> {\rangle }
\def \^ {\hat}

\newcommand{\norm}[1]{\left \|#1\right \|}

\newcommand{\inftynorm}[1]{\norm{#1}_\infty}

\newcommand{\fronorm}[1]{\norm{#1}_F}
\newcommand{\nucnorm}[1]{\norm{#1}_*}

\newcommand{\abs}[1]{\left | #1 \right |}

\newcommand{\mat}[1]{\boldsymbol{#1}}

\def \S {\mat{S}}

\def \Y {\mat{Y}}

\def \loglike {\ell}

\newcommand{\ind}[1]{\mathds{1}_{[#1]}}

\definecolor{noteColor}{RGB}{200,50,50}

\begin{document}

\title{Misclassification excess risk bounds for 1-bit matrix completion}

\author[1]{The Tien Mai}

\authormark{MAI \textsc{et al.}}
\titlemark{Misclassification excess risk bounds for 1-bit matrix completion}

\address{\orgdiv{Department of Mathematical Sciences}, \orgname{Norwegian University of Science and Technology}, \orgaddress{\state{Trondheim 7034}, \country{Norway}}}

\corres{\email{the.t.mai@ntnu.no}}



\abstract[Abstract]{
This study investigates the misclassification excess risk bound in the context of 1-bit matrix completion, a significant problem in machine learning involving the recovery of an unknown matrix from a limited subset of its entries. Matrix completion has garnered considerable attention in the last two decades due to its diverse applications across various fields. Unlike conventional approaches that deal with real-valued samples, 1-bit matrix completion is concerned with binary observations. While prior research has predominantly focused on the estimation error of proposed estimators, our study shifts attention to the prediction error. This paper offers theoretical analysis regarding the prediction errors of two previous works utilizing the logistic regression model: one employing a max-norm constrained minimization and the other employing nuclear-norm penalization. Significantly, our findings demonstrate that the latter achieves the minimax-optimal rate without the need for an additional logarithmic term. These novel results contribute to a deeper understanding of 1-bit matrix completion by shedding light on the predictive performance of specific methodologies.
}


\jnlcitation{\cname{%
\author{Mai T.T.
}.}
\ctitle{Misclassification excess risk bounds for 1-bit matrix completion.} 
}

\maketitle



\section{Introduction}

The matrix completion problem is an extensively investigated issue within the realms of machine learning and statistics, garnering significant attention in recent years. This surge in interest is fueled by contemporary applications such as recommendation systems \cite{bobadilla2013recommender,koren2009matrix} and particularly the famous Netflix challenge \cite{bennett2007netflix},  image processing \cite{ji2010robust,han2014linear},  genotype imputation \cite{chi2013genotype,jiang2016sparrec},  and quantum statistics \cite{gross2011recovering}. The task of reconstructing a matrix without any supplementary information is inherently considered unattainable. Nevertheless, the feasibility of this task may be enhanced under specific assumptions regarding the inherent structure of the matrix awaiting recovery. This feasibility is demonstrated by works such as \cite{CandesT10,candes2010matrix,CandesRecht}, where the pivotal assumption is that the matrix has a low rank structure. This assumption is particularly natural in various practical scenarios, such as recommendation systems, where it signifies the presence of a small number of latent features elucidating user preferences. Different approaches for matrix completion have been proposed and studied from theoretical and computational points of views, see for example \cite{tsybakov2011nuclear,lim2007variational,salakhutdinov2008bayesian,recht2013parallel,chatterjee2015matrix,mai2015,alquier2020concentration,chen2019inference}.

The previously mentioned papers primarily concentrated on matrices with real-valued entries. Nevertheless, in numerous practical applications, the observed entries are subjected to significant quantization, frequently restricted to a single bit and belonging to the set $\{-1, 1\}$. This quantization scheme is prevalent in various scenarios, such as voting or rating data and survey responses. In these contexts, the typical nature of responses involves binary distinctions, such as "yes/no", "like/dislike", or "true/false", highlighting the inherent discrete and binary characteristics of the observed data.
The problem
of recovering a  matrix from partial binary (1-bit) observations is usually referred to
as 1-bit matrix completion which was first introduced and studied in \cite{Davenport14}. Since then, various works have been proposed and studied in the context of 1-bit matrix completion, see for examples \cite{cai2013,klopp2015adaptive,hsieh2015pu,cottet20181bit,herbster2016mistake,alquier2019estimation} among others. The predominant emphasis in existing studies lies in addressing estimation errors, demonstrating the consistency of the proposed estimators. However, there is a notable scarcity of attention dedicated to ensuring a guarantee on the probability of encountering prediction errors.

From a machine learning point of view, dealing 
with binary output is called a classification problem which is one of the most important setups in statistical learning. This problem has been
studied in various contexts, see for examples in \cite{devroye2013probabilistic,vapnik1999nature}, while the
surveys of the state-of-the-art can be found in \cite{boucheron2005theory,hastie2009elements,buhlmann2011statistics,giraud2021introduction}. Let's consider a binary classification scenario involving a high-dimensional feature vector denoted as $x \in \mathbb{R}^d$ and the class label outcome $Y|x $ follows a binomial distribution with parameters $ p(x) $, as in \cite{abramovich2018high}.
The accuracy of a classifier $\eta$
is defined by a misclassification risk 
$$
R(\eta)
=
\mathbb{P} (Y \neq \eta(x)).
$$ 
It is well-known that the Bayes classifier,
$$
\eta^*(x)
=
1\{ p(x) \geq 1/2 \}
,
$$
minimizes $R(\eta) $, i.e $ \eta^* = \argmin R(\eta)$.
However, the probability function $p(x)$ is unknown and the resulting classifier $\hat{\eta}(x)$ should be designed from the data $ \mathcal{S} $: a random sample of $n$ independent observations $(x_1,Y_1),\ldots, (x_n,Y_n)$. 
The design points $x_i$ may be considered as fixed or random.  
The corresponding (conditional) misclassification error of $\hat{\eta}$ is 
$$
R(\hat{\eta})
=
\mathbb{P} (Y \neq \hat{\eta}(x) \, | \, \mathcal{S} )
$$ 
and the goodness of $\hat{\eta} $ with respect to $\eta^*$ is measured by the misclassification
excess risk, defined as
$$
excess(\hat{\eta},\eta^*)
=
\mathbb{E} R(\hat{\eta})-R(\eta^*).
$$

A standard approach to obtain $ \hat{\eta}$ is to assume some (parametric or nonparametric) model for $p( x)$. The most commonly used models is logistic regression, where it is assumed that
$p( x)= e^{\beta^t  x} /(1+ e^{\beta^t  x} ) $ and $\beta \in \mathbb{R}^d$ is a vector of unknown regression coefficients. The corresponding Bayes classifier 
is a linear classifier $\eta^*( x)= 1\{p( x) \geq 1/2\}= 1\{\beta^t  x \geq 0\}$.
One then estimates $\beta$ from the data by the maximum likelihood estimator $ \hat{\beta}$,
plugs-in $ \hat{\beta}$ and
the resulting (linear) classifier is $ \hat{\eta}( x)= 1\{ \hat{p}( x) \geq 1/2\}= 1\{ \hat{\beta}^t  x \geq 0\}$. This approach has been adapted to the context of 1-bit matrix completion first in the work \cite{Davenport14} and then studied in \cite{cai2013, klopp2015adaptive,alaya2019collective,hsieh2015pu}.

An alternative (nonparametric) technique for deriving a classifier, $\hat{\eta}$, from available data is empirical risk minimization. In this approach, the objective is to minimize the empirical risk, $ \hat{R}_n(\eta) $, which is a substitute for the true misclassification risk $R(\eta)$.
$$
\hat{R}_n(\eta)
=
\frac{1}{n}\sum_{i=1}^n 1\{Y_i \neq \eta(x_i)\}
$$
over a given class of classifiers, see for examples \cite{boucheron2005theory,giraud2021introduction}.
However, applying this strategy directly in practical applications poses computational challenges due to its non-convex and non-smooth nature. Typically, this challenge is mitigated by adopting a related convex minimization surrogate, such as hinge loss \cite{zhang2004statistical,bartlett2006convexity}. Notably, this approach has been less explored in the context of 1-bit matrix completion. To the best of our knowledge, the initial efforts to apply this method to 1-bit matrix completion are found in the works of \cite{cottet20181bit} and \cite{alquier2019estimation}.

\subsection*{Related works and contributions}

In the context of 1-bit matrix completion, the observation of \(n\) samples \(Y_{ij}\), where \(1 \leq i \leq m_1\) and \(1 \leq j \leq m_2\), follows a Bernoulli distribution with parameter \(f(M_{ij})\). Here, \(f\) is a link function mapping from \(\mathbb{R}\) to the interval \([0, 1]\), and \(M\) is a \(m_1 \times m_2\) real matrix to be recovered, as detailed in references \cite{cai2013,Davenport14,klopp2015adaptive}.

While substantial progress has been made in addressing the estimation error for 1-bit matrix completion, as discussed in Section \ref{sc_on_estimation}, there is a noticeable scarcity of studies focusing on the misclassification excess risk. The pioneering work in addressing misclassification excess risk, to the best of our knowledge, is presented in \cite{cottet20181bit}, which studies a variational Bayesian method with hinge loss. However, the study of misclassification excess risk is confined to a highly restrictive noiseless setting and utilizes PAC-Bayesian techniques.
An alternative approach is proposed in \cite{alquier2019estimation}, which introduces a nuclear-norm penalization method and achieves a similar result, using hinge loss, to \cite{cottet20181bit} in a more general case. Specifically, they establish a misclassification excess error of order \(r^*(m_1 + m_2) \log (m_1 + m_2)/n\), where \(r^*\) represents the rank of \(M^*\), and \(M^*\) is the true model parameter.

For logistic models, \cite{alquier2019estimation} obtains a slower rate of \(\sqrt{r^*(m_1 + m_2) \log (m_1 + m_2)/n}\), and similarly, \cite{alaya2019collective} derives a misclassification excess error of order \(\sqrt{r^*(m_1 + m_2)/n}\) without an additional logarithmic term. 

In this work, we make two significant contributions: firstly, it establishes an upper bound for misclassification excess error for the method proposed by \cite{cai2013}, and secondly, it demonstrates that for logistic models, a fast rate of misclassification excess error of order \( r^*(m_1 + m_2)/n \) can be achieved without an extra logarithmic term, which is minimax-optimal. These contributions are novel as of our current knowledge.

\subsection*{Notation and organization of the paper}

In this paper, we use $[d]$ to denote the set of integers $\{1, \ldots, d\}$.  We use capital letter to denote a matrix (e.g., $ M $) and standard text to denote its entries (e.g., $M_{i,j}$). We let $\|M\| $ denote the operator norm of $M $, $\fronorm{M} = \sqrt{\sum_{i,j} M_{i,j}^2}$ denote the Frobenius norm of $M$, $\nucnorm{M}$ denote the nuclear or Schatten-1 norm of $M$ (the sum of the singular values), and  $\inftynorm{ M} = \max_{i,j} \abs{M_{i,j}}$ denote the entry-wise infinity-norm of $M$ and for $ p,q \geq 1 $ put $ \|M\|_{p,q} = (\sum_j (\sum_i |M_ij|^p )^{q/p})^{1/q} $. 
Finally, for an event $\mathcal{E}$ ,$\ind{\mathcal{E}}$ is the indicator function for that event, i.e., $\ind{\mathcal{E}}$ is $1$ if $\mathcal{E}$ occurs and $0$ otherwise. The matrix max-norm of $ M $ is defined as $ \|M\|_{max} = \min_{M= UV^\top} \|U\|_{2,\infty} \|V\|_{2,\infty} $.

The rest of the paper is organized as follows: In Section \ref{sec_model}, we outline the 1-bit matrix completion problem, introduce various methods to address it, and provide a brief discussion on estimation error. Section \ref{sc_misclass_error} presents our primary results concerning misclassification excess error across different methods. All technical proofs are given in Section \ref{sc_proofs}. We summarize our findings in the conclusion, covered in Section \ref{sc_conclusion}.

\section{Model and Method}
\label{sec_model}
\subsection{One-bit matrix completion}
\label{subsec:bin}
Assume that the observations follow a binomial distribution
parametrized by a matrix $  X^* \in \mathbb{R}^{m_1\times m_2}$.
Assume in addition that an $i.i.d $ sequence of coefficients (indexes) $(\omega_i)_{i=1}^n \in ([m_1] \times [m_2])^n, n < m_1 m_2, $
is revealed uniformly, put $ \Omega = (\omega_i)_{i=1}^n $.
The observations associated to these coefficients are denoted by $
(Y_i)_{i=1}^n \in \{-1,1\}^n
$
and distributed as follows
\begin{equation}
\label{eq_observations}
Y_{i} 
= 
\begin{cases} +1 & \mathrm{with~probability} \ f( X_{\omega_i}), \\
-1 & \mathrm{with~probability} \ 1- f( X_{\omega_i}) 
.
\end{cases}
\end{equation}
where $f $ is the logistic link function, which is common in statistics, with $f(x) = e^x/(1 + e^x) $.
For ease of notation, we often write $ X_i$  instead of  $ X_{\omega_i}$.  Specifically, we assume that each entry of $ [m_1] \times [m_2]\, , \omega_i, $ is uniformly observed with probability $ n/(m_1 m_2) $, independently.

Define
\begin{equation*}
\label{eq:definition-d-M}
d= m_1+ m_2 , 
\quad 
M= \max (m_1 , m_2), 
\quad 
m= \min (m_1 , m_2)  .
\end{equation*}

In order to estimate either $ X^* $ or $f(X^*)$, a first strategy was propose in \cite{Davenport14} that maximizes the log-likelihood function of the optimization variable $ X$ given the observations subject to a set of convex constraints.  In our case, the log-likelihood function is given by
\[
\loglike_{\Omega, Y}( X) 
:= 
\sum_{(i,j) \in \Omega}  
\left(
\ind{Y_{i,j} = 1} \log(f(X_{i,j})) + \ind{Y_{i,j} = -1} \log(1 - f(X_{i,j})) 
\right).
\]

Put, for some $ r, \gamma >0 $. 
\begin{align*}
K_{*} (\gamma,r) 
&= 
\{ X \in \mathbb{R}^{m_1\times m_2 } : 
\|X\|_\infty \leq \gamma , \|X\|_* \leq \gamma \sqrt{r\, m_1 m_2} \},
\\
K_{max} (\gamma,r) 
& = 
\{ X \in \mathbb{R}^{m_1\times m_2 } : 
\|X\|_\infty \leq \gamma , \|X\|_{max} \leq \gamma \sqrt{r} \}.
\end{align*}
It is noted that for matrices $ X $ of rank at most $ r $ we have that $ K_{max} (\gamma,r) \subset  K_{*} (\gamma,r)  $, see \cite{cai2013}.

The reference \cite{Davenport14} employs the following convex program:
\begin{equation}
\label{eq:max likelihood}
\hat{ X}^{(1)} = \argmax_{ X\in K_{*} (\gamma,r) }  \loglike_{\Omega, \Y}( X).
\end{equation}
Additionally, \cite{cai2013} proposes a max-norm constrained method as:
\begin{equation}
\label{eq_maxnrom likelihood}
\hat{ X}^{(2)} = \argmax_{ X\in K_{max} (\gamma,r) }  \loglike_{\Omega, \Y}( X).
\end{equation}

Instead of adopting a constrained method, \cite{klopp2015adaptive} suggests a nuclear-norm penalized approach as follows:
\begin{equation}
\label{eq_nu_pen_estimator}
\hat{ X}^{nu-pen}
=
\argmin_{\substack{\| X\|_{\infty}\leq \gamma }}
\Phi_Y^\lambda 
( X), \,
\text{where} 
\quad  
\Phi_Y^\lambda ( X)
= 
\frac{1}{n}\loglike_{\Omega, Y}( X) +
\lambda  \| X \|_{*} \,
\end{equation}
with $\lambda>0$ being a regularization parameter. This estimator undergoes further examination in \cite{alquier2019estimation} and is expanded to a broader context in \cite{alaya2019collective}.

\subsection{On the estimation error}
\label{sc_on_estimation}
Under the assumption that $ {\rm rank}(X^*) \leq r $ and $ \| X\|_{\infty}\leq \gamma $ the authors of \cite{Davenport14} show that 
$$
\frac{1}{m_1 m_2} \|\hat{ X}^{(1)} -  X^* \|_{F}^2
\leq 
c_\gamma \sqrt{\frac{rd}{n }} 
$$
where $ c_\gamma $ is a constant depending only on $ \gamma $, see Theorem 1 in \cite{Davenport14}. A similar
result using max-norm minimization was also obtained in \cite{cai2013}. The paper \cite{klopp2015adaptive} proves a faster estimation error rate as  
$$
\frac{1}{m_1 m_2} \|\hat{ X}^{nu-pen} -  X^* \|_{F}^2
\leq 
c_\gamma \frac{rd\log(d)}{n}
,
$$
A comparable result is established in \cite{alquier2019estimation}, specifically in Theorem 4.2. Subsequently, this rate has been recently enhanced to \( rd/n \), without the presence of a logarithmic term, as demonstrated in \cite{alaya2019collective} (refer to Theorem 7). Moreover, Theorem 3 in \cite{klopp2015adaptive} establishes a lower bound, indicating that the estimation error cannot surpass \( rd/n \). Consequently, the work presented in \cite{alaya2019collective} attains the precise minimax estimation rate of convergence for 1-bit matrix completion.

\section{Misclassification excess risk bounds}	
\label{sc_misclass_error}

Following the standard approach in classification
theory~\cite{vapnik1999nature}, we now study the misclassification excess risk for 1-bit matrix completion.
The corresponding (ideal) Bayes classifier, in the logistic regression, 
is a linear classifier 
\begin{align*}
\eta^*(X^*_\omega)
& =
\begin{cases} 
+1 & \mathrm{if} \ f( X^*_{\omega}) \geq 0.5, 
\\
-1 & \mathrm{if} \ f( X^*_{\omega}) < 0.5,
.
\end{cases}
\\
& =
\sign(  X^*_{\omega}).
\end{align*}
The accuracy of a classifier $ \eta $ to predict a new entry of the matrix is
assessed by a misclassification error
\begin{align*}
R( X) 
= 
\mathbb{P}\left[ Y \neq \eta ( X_\omega) \right].
\end{align*}

One then estimates $ X $ from the data $ \mathcal{S} $ consisting of a random sample of $ n $ independent observations $ (Y_1,\omega_1), \ldots , (Y_n,\omega_n) $ by using one of the methods mentioned above to get an estimator $ \hat{ X} $. 
The corresponding (conditional) misclassification error of $ \hat{\eta} $ is 
$$
R(\hat{\eta})
=
\mathbb{P} (Y \neq \hat{\eta}(\hat{ X}) \, | \, \mathcal{S} )
$$ 
and the goodness of $\hat{\eta}$ w.r.t. $\eta^*$ is measured by the misclassification
excess risk 
$$
excess(\hat{\eta},\eta^*)
=
\mathbb{E} \, R(\hat{\eta}) - \inf_\eta R(\eta)
=
\mathbb{E} \, R(\hat{\eta})-R(\eta^*) .
$$

\subsection{Misclassification excess risk bound}
We now present a first result on misclassification excess error for the method proposed in \cite{cai2013}.
The proof is given in Section \ref{sc_proofs}.

\begin{theorem}
	\label{thm_slowrate}
	We have 
	for the estimator $ \hat{ X}^{(2)} $ that
	\begin{align*}
	excess(\hat{\eta},\eta^*)
	\leq 
	\sqrt{ 
		C_\gamma \sqrt{\frac{\rank(X^*)d}{n }} 
	}
	\end{align*}
	where $ C_\gamma $ is a constant depending only on $ \gamma $.
\end{theorem}

Up to our knowledge, the misclassification excess risk bound in Theorem \ref{thm_slowrate} is novel. It brings more information on the behavior of the methods proposed in \cite{cai2013} and\cite{Davenport14}.

\begin{remark}
In the logistic regression model, the rate as established in Theorem \ref{thm_slowrate} is comparatively slower than the one derived by \cite{alaya2019collective} (refer to Corollary 15, page 13), which is of the order \( \sqrt{\text{rank}(X^*)d/n} \). However, it is crucial to acknowledge that the result in \cite{alaya2019collective} is obtained under the additional assumption of the so-called 'low-noise' condition. Importantly, in the subsequent section, we demonstrate that the method presented in \cite{cai2013} can achieve the same rate as in \cite{alaya2019collective} when this type of assumption is satisfied.
\end{remark}

\subsection{Improved bounds with low-noise assumption}

The primary challenges for any classifier manifest in the vicinity of the boundary \(\{x: p(x)=1/2\}\), or equivalently, a hyperplane \(\beta^t x=0\) for the logistic regression model, where accurate prediction of the class label becomes particularly challenging. However, in regions where \(p(x)\) is sufficiently away from \(1/2\) (referred to as the margin or low-noise condition), there exists potential for improving the bounds on misclassification excess risk, as established in the preceding section.

In alignment with the approach outlined in \cite{bartlett2006convexity} (see Lemma 5, page 146), we introduce the following low-noise assumption:
\begin{assumption}[\bf LN] 
	\label{as:B}   
	Consider the logistic regression model (\ref{eq_observations}) and assume that, for some $ c>0 $,
	\begin{equation}
	\label{eq:asb}
	\mathbb{P} \left( 0 < \bigg| f(X) - \frac{1}{2} \bigg| < \frac{1}{2c} \right)  = 0.
	\end{equation}
\end{assumption}
Assumption (LN) essentially assumes the existence of the probability is not very close to 1/2. This kind of assumption was introduced in \cite{mammen1999smooth,tsybakov2004optimal}.

\begin{theorem}
	\label{thrm_slow_margin}
Under the assumption (LN).
We have 
	for the estimator $ \hat{ X}^{(2)} $ that
	\begin{align*}
	excess(\hat{\eta},\eta^*)
	\leq 
	C_\gamma \sqrt{\frac{\rank(X^*)d}{n }} 
	\end{align*}
	where $ C_\gamma $ is a constant depending only on $ \gamma $.
\end{theorem}
\begin{remark}
The upper bound for misclassification excess risk, as presented in Theorem \ref{thrm_slow_margin}, shares similarities with the results reported in \cite{alquier2019estimation} (Theorem 4.2) and \cite{alaya2019collective} (Corollary 15 and Remark 9). Notably, our technical approach is considerably simpler compared to the methodologies employed in these two references.
\end{remark}

\begin{remark}
A faster rate of \( \text{rank}(X^*)M \log(d)/n \) for classification excess error was initially achieved in \cite{cottet20181bit}, to the best of our knowledge. However, it is important to note that this result is obtained under a highly restrictive noiseless setting. Furthermore, the study in \cite{cottet20181bit} employs a variational Bayesian method using hinge loss and utilizes the PAC-Bayesian bound technique.

A comparable rate, \( \text{rank}(X^*)M \log(d)/n \), is also obtained by an alternative estimator based on the hinge loss, rather than the logistic loss, as demonstrated in Theorem 4.4 on page 2319 of \cite{alquier2019estimation}. However, it is noteworthy that the analysis in \cite{alquier2019estimation} for the constraint nuclear norm estimator with logistic loss yields a slower rate. Specifically, Theorem 4.2 in \cite{alquier2019estimation} (pages 2137-2138) establishes that the rate for the excess is \( \sqrt{\text{rank}(X^*)M \log(d)/n} \).
\end{remark}

To the best of our knowledge, the misclassification excess risk bound presented in Theorem \ref{thrm_slow_margin} is also novel. Despite its comparatively slower rate compared to \cite{cottet20181bit} and \cite{alquier2019estimation}, this result contributes valuable insights into the performance characteristics of the methods introduced in \cite{cai2013} and \cite{Davenport14}.

\subsection{Sharp rate}
In the subsequent theorem, we demonstrate that a more precise misclassification excess error, without an additional logarithmic term, can be achieved for the nuclear-norm penalization method within the logistic model. This outcome signifies an enhancement over the findings presented in the reference \cite{alaya2019collective}.
\begin{theorem}
	\label{thrm_nuclear_margin_sharprate}
Under the assumption (LN),
	for the estimator $ \hat{ X}^{nu-pen} $, we have that
	\begin{align*}
	excess(\hat{\eta},\eta^*)
	\leq 
	C_\gamma \frac{\rank(X^*)d}{n},
	\end{align*}
	where $ C_\gamma $ is a constant depending only on $ \gamma $.
\end{theorem}

The distinctive technical approach employed to derive the results in the aforementioned theorem lies in leveraging the low-noise assumption. By utilizing the outcomes from the paper \cite{bartlett2006convexity}, a faster rate is achieved, deviating from the methodology in \cite{alaya2019collective} where the results from \cite{zhang2004statistical} were employed. This divergence in approach enables the attainment of a more rapid rate.

\begin{remark}
In Theorem 4.5 of \cite{alquier2019estimation} on page 2140, a lower bound of the order \( \text{rank}(X^*)d/n \) for the misclassification excess risk was established. Consequently, Theorem \ref{thrm_nuclear_margin_sharprate} demonstrates that the estimator \( \hat{ X}^{nu-pen} \) attains the optimal-minimax misclassification excess rate without the presence of a logarithmic term. This constitutes a novel contribution that enhances our understanding of 1-bit matrix completion.
\end{remark}

\section{Conclusion}
\label{sc_conclusion}
In this work, we investigated the problem of matrix completion, which involves the recovery of a matrix from an incomplete set of sampled entries. Specifically, our focus was on binary observations, a well-suited context for example for voting data expressed as "yes/no" or "true/false" responses. Our analysis delves into the misclassification excess error associated with certain methods designed for 1-bit matrix completion—a topic that has received limited attention in existing literature.

Our study presents significant original contributions. Firstly, we establish an upper bound for the misclassification excess error concerning the method proposed in \cite{cai2013}. Secondly, we demonstrate that a method incorporating nuclear norm penalization with a logistic regression model can achieve a fast rate of convergence. Furthermore, we establish that this rate is minimax-optimal without additional (multiplicative) logarithmic terms.

It is important to highlight that the current analysis, as well as in \cite{Davenport14,cai2013,klopp2015adaptive,alaya2019collective}, assumes the correctness of the model, specifically that the link function \(f\) is accurately specified. A potential avenue for future research could involve exploring scenarios where model mis-specification occurs and investigating the associated prediction error. While some initial work in this direction has been undertaken without model specification in \cite{cottet20181bit}, to the best of our knowledge, such exploration has not been pursued for nuclear-norm penalization or constrained methods.

\bmsection*{Author contributions}
I am the only author of this paper.

\bmsection*{Acknowledgments}
TTM is supported by the Norwegian Research Council, grant number 309960 through the Centre for Geophysical Forecasting at NTNU.

\bmsection*{Financial disclosure}

None reported.

\bmsection*{Conflict of interest}

The authors declare no potential conflict of interests.


\begin{thebibliography}{}
	
	\bibitem [\protect \citeauthoryear {%
		Abramovich%
		\ \BBA {} Grinshtein%
	}{%
		Abramovich%
		\ \BBA {} Grinshtein%
	}{%
		{\protect \APACyear {2018}}%
	}]{%
		abramovich2018high}
	\APACinsertmetastar {%
		abramovich2018high}%
	\begin{APACrefauthors}%
		Abramovich, F.%
		\BCBT {}\ \BBA {} Grinshtein, V.%
	\end{APACrefauthors}%
	\unskip\
	\newblock
	\APACrefYearMonthDay{2018}{}{}.
	\newblock
	{\BBOQ}\APACrefatitle {High-dimensional classification by sparse logistic
		regression} {High-dimensional classification by sparse logistic
		regression}.{\BBCQ}
	\newblock
	\APACjournalVolNumPages{IEEE Transactions on Information
		Theory}{65}{5}{3068--3079}.
	\PrintBackRefs{\CurrentBib}
	
	\bibitem [\protect \citeauthoryear {%
		Alaya%
		\ \BBA {} Klopp%
	}{%
		Alaya%
		\ \BBA {} Klopp%
	}{%
		{\protect \APACyear {2019}}%
	}]{%
		alaya2019collective}
	\APACinsertmetastar {%
		alaya2019collective}%
	\begin{APACrefauthors}%
		Alaya, M\BPBI Z.%
		\BCBT {}\ \BBA {} Klopp, O.%
	\end{APACrefauthors}%
	\unskip\
	\newblock
	\APACrefYearMonthDay{2019}{}{}.
	\newblock
	{\BBOQ}\APACrefatitle {Collective Matrix Completion.} {Collective matrix
		completion.}{\BBCQ}
	\newblock
	\APACjournalVolNumPages{J. Mach. Learn. Res.}{20}{}{148--1}.
	\PrintBackRefs{\CurrentBib}
	
	\bibitem [\protect \citeauthoryear {%
		Alquier%
		, Cottet%
		\BCBL {}\ \BBA {} Lecu{\'e}%
	}{%
		Alquier%
		\ \protect \BOthers {.}}{%
		{\protect \APACyear {2019}}%
	}]{%
		alquier2019estimation}
	\APACinsertmetastar {%
		alquier2019estimation}%
	\begin{APACrefauthors}%
		Alquier, P.%
		, Cottet, V.%
		\BCBL {}\ \BBA {} Lecu{\'e}, G.%
	\end{APACrefauthors}%
	\unskip\
	\newblock
	\APACrefYearMonthDay{2019}{}{}.
	\newblock
	{\BBOQ}\APACrefatitle {Estimation bounds and sharp oracle inequalities of
		regularized procedures with Lipschitz loss functions} {Estimation bounds and
		sharp oracle inequalities of regularized procedures with lipschitz loss
		functions}.{\BBCQ}
	\newblock
	\APACjournalVolNumPages{The Annals of Statistics}{47}{4}{2117--2144}.
	\PrintBackRefs{\CurrentBib}
	
	\bibitem [\protect \citeauthoryear {%
		Alquier%
		\ \BBA {} Ridgway%
	}{%
		Alquier%
		\ \BBA {} Ridgway%
	}{%
		{\protect \APACyear {2020}}%
	}]{%
		alquier2020concentration}
	\APACinsertmetastar {%
		alquier2020concentration}%
	\begin{APACrefauthors}%
		Alquier, P.%
		\BCBT {}\ \BBA {} Ridgway, J.%
	\end{APACrefauthors}%
	\unskip\
	\newblock
	\APACrefYearMonthDay{2020}{}{}.
	\newblock
	{\BBOQ}\APACrefatitle {Concentration of tempered posteriors and of their
		variational approximations} {Concentration of tempered posteriors and of
		their variational approximations}.{\BBCQ}
	\newblock
	\APACjournalVolNumPages{The Annals of Statistics}{48}{3}{1475--1497}.
	\PrintBackRefs{\CurrentBib}
	
	\bibitem [\protect \citeauthoryear {%
		Bartlett%
		, Jordan%
		\BCBL {}\ \BBA {} McAuliffe%
	}{%
		Bartlett%
		\ \protect \BOthers {.}}{%
		{\protect \APACyear {2006}}%
	}]{%
		bartlett2006convexity}
	\APACinsertmetastar {%
		bartlett2006convexity}%
	\begin{APACrefauthors}%
		Bartlett, P\BPBI L.%
		, Jordan, M\BPBI I.%
		\BCBL {}\ \BBA {} McAuliffe, J\BPBI D.%
	\end{APACrefauthors}%
	\unskip\
	\newblock
	\APACrefYearMonthDay{2006}{}{}.
	\newblock
	{\BBOQ}\APACrefatitle {Convexity, classification, and risk bounds} {Convexity,
		classification, and risk bounds}.{\BBCQ}
	\newblock
	\APACjournalVolNumPages{Journal of the American Statistical
		Association}{101}{473}{138--156}.
	\PrintBackRefs{\CurrentBib}
	
	\bibitem [\protect \citeauthoryear {%
		Bennett%
		\ \BBA {} Lanning%
	}{%
		Bennett%
		\ \BBA {} Lanning%
	}{%
		{\protect \APACyear {2007}}%
	}]{%
		bennett2007netflix}
	\APACinsertmetastar {%
		bennett2007netflix}%
	\begin{APACrefauthors}%
		Bennett, J.%
		\BCBT {}\ \BBA {} Lanning, S.%
	\end{APACrefauthors}%
	\unskip\
	\newblock
	\APACrefYearMonthDay{2007}{}{}.
	\newblock
	{\BBOQ}\APACrefatitle {The netflix prize} {The netflix prize}.{\BBCQ}
	\newblock
	\BIn{} \APACrefbtitle {Proceedings of KDD cup and workshop} {Proceedings of kdd
		cup and workshop}\ (\BVOL\ 2007, \BPG~35).
	\PrintBackRefs{\CurrentBib}
	
	\bibitem [\protect \citeauthoryear {%
		Bobadilla%
		, Ortega%
		, Hernando%
		\BCBL {}\ \BBA {} Guti{\'e}rrez%
	}{%
		Bobadilla%
		\ \protect \BOthers {.}}{%
		{\protect \APACyear {2013}}%
	}]{%
		bobadilla2013recommender}
	\APACinsertmetastar {%
		bobadilla2013recommender}%
	\begin{APACrefauthors}%
		Bobadilla, J.%
		, Ortega, F.%
		, Hernando, A.%
		\BCBL {}\ \BBA {} Guti{\'e}rrez, A.%
	\end{APACrefauthors}%
	\unskip\
	\newblock
	\APACrefYearMonthDay{2013}{}{}.
	\newblock
	{\BBOQ}\APACrefatitle {Recommender systems survey} {Recommender systems
		survey}.{\BBCQ}
	\newblock
	\APACjournalVolNumPages{Knowledge-based systems}{46}{}{109--132}.
	\PrintBackRefs{\CurrentBib}
	
	\bibitem [\protect \citeauthoryear {%
		Boucheron%
		, Bousquet%
		\BCBL {}\ \BBA {} Lugosi%
	}{%
		Boucheron%
		\ \protect \BOthers {.}}{%
		{\protect \APACyear {2005}}%
	}]{%
		boucheron2005theory}
	\APACinsertmetastar {%
		boucheron2005theory}%
	\begin{APACrefauthors}%
		Boucheron, S.%
		, Bousquet, O.%
		\BCBL {}\ \BBA {} Lugosi, G.%
	\end{APACrefauthors}%
	\unskip\
	\newblock
	\APACrefYearMonthDay{2005}{}{}.
	\newblock
	{\BBOQ}\APACrefatitle {Theory of classification: A survey of some recent
		advances} {Theory of classification: A survey of some recent
		advances}.{\BBCQ}
	\newblock
	\APACjournalVolNumPages{ESAIM: probability and statistics}{9}{}{323--375}.
	\PrintBackRefs{\CurrentBib}
	
	\bibitem [\protect \citeauthoryear {%
		B{\"u}hlmann%
		\ \BBA {} Van De~Geer%
	}{%
		B{\"u}hlmann%
		\ \BBA {} Van De~Geer%
	}{%
		{\protect \APACyear {2011}}%
	}]{%
		buhlmann2011statistics}
	\APACinsertmetastar {%
		buhlmann2011statistics}%
	\begin{APACrefauthors}%
		B{\"u}hlmann, P.%
		\BCBT {}\ \BBA {} Van De~Geer, S.%
	\end{APACrefauthors}%
	\unskip\
	\newblock
	\APACrefYear{2011}.
	\newblock
	\APACrefbtitle {Statistics for high-dimensional data: methods, theory and
		applications} {Statistics for high-dimensional data: methods, theory and
		applications}.
	\newblock
	\APACaddressPublisher{}{Springer Science \& Business Media}.
	\PrintBackRefs{\CurrentBib}
	
	\bibitem [\protect \citeauthoryear {%
		Cai%
		\ \BBA {} Zhou%
	}{%
		Cai%
		\ \BBA {} Zhou%
	}{%
		{\protect \APACyear {2013}}%
	}]{%
		cai2013}
	\APACinsertmetastar {%
		cai2013}%
	\begin{APACrefauthors}%
		Cai, T.%
		\BCBT {}\ \BBA {} Zhou, W\BHBI X.%
	\end{APACrefauthors}%
	\unskip\
	\newblock
	\APACrefYearMonthDay{2013}{}{}.
	\newblock
	{\BBOQ}\APACrefatitle {A max-norm constrained minimization approach to 1-bit
		matrix completion.} {A max-norm constrained minimization approach to 1-bit
		matrix completion.}{\BBCQ}
	\newblock
	\APACjournalVolNumPages{J. Mach. Learn. Res.}{14}{1}{3619--3647}.
	\PrintBackRefs{\CurrentBib}
	
	\bibitem [\protect \citeauthoryear {%
		Candes%
		\ \BBA {} Plan%
	}{%
		Candes%
		\ \BBA {} Plan%
	}{%
		{\protect \APACyear {2010}}%
	}]{%
		candes2010matrix}
	\APACinsertmetastar {%
		candes2010matrix}%
	\begin{APACrefauthors}%
		Candes, E\BPBI J.%
		\BCBT {}\ \BBA {} Plan, Y.%
	\end{APACrefauthors}%
	\unskip\
	\newblock
	\APACrefYearMonthDay{2010}{}{}.
	\newblock
	{\BBOQ}\APACrefatitle {Matrix completion with noise} {Matrix completion with
		noise}.{\BBCQ}
	\newblock
	\APACjournalVolNumPages{Proceedings of the IEEE}{98}{6}{925--936}.
	\PrintBackRefs{\CurrentBib}
	
	\bibitem [\protect \citeauthoryear {%
		Cand{\`e}s%
		\ \BBA {} Recht%
	}{%
		Cand{\`e}s%
		\ \BBA {} Recht%
	}{%
		{\protect \APACyear {2009}}%
	}]{%
		CandesRecht}
	\APACinsertmetastar {%
		CandesRecht}%
	\begin{APACrefauthors}%
		Cand{\`e}s, E\BPBI J.%
		\BCBT {}\ \BBA {} Recht, B.%
	\end{APACrefauthors}%
	\unskip\
	\newblock
	\APACrefYearMonthDay{2009}{}{}.
	\newblock
	{\BBOQ}\APACrefatitle {Exact matrix completion via convex optimization} {Exact
		matrix completion via convex optimization}.{\BBCQ}
	\newblock
	\APACjournalVolNumPages{Found. Comput. Math.}{9}{6}{717--772}.
	\newblock
	\begin{APACrefDOI} 10.1007/s10208-009-9045-5 \end{APACrefDOI}
	\PrintBackRefs{\CurrentBib}
	
	\bibitem [\protect \citeauthoryear {%
		Cand{\`e}s%
		\ \BBA {} Tao%
	}{%
		Cand{\`e}s%
		\ \BBA {} Tao%
	}{%
		{\protect \APACyear {2010}}%
	}]{%
		CandesT10}
	\APACinsertmetastar {%
		CandesT10}%
	\begin{APACrefauthors}%
		Cand{\`e}s, E\BPBI J.%
		\BCBT {}\ \BBA {} Tao, T.%
	\end{APACrefauthors}%
	\unskip\
	\newblock
	\APACrefYearMonthDay{2010}{}{}.
	\newblock
	{\BBOQ}\APACrefatitle {The power of convex relaxation: near-optimal matrix
		completion} {The power of convex relaxation: near-optimal matrix
		completion}.{\BBCQ}
	\newblock
	\APACjournalVolNumPages{IEEE Trans. Inform. Theory}{56}{5}{2053--2080}.
	\newblock
	\begin{APACrefDOI} 10.1109/TIT.2010.2044061 \end{APACrefDOI}
	\PrintBackRefs{\CurrentBib}
	
	\bibitem [\protect \citeauthoryear {%
		Chatterjee%
	}{%
		Chatterjee%
	}{%
		{\protect \APACyear {2015}}%
	}]{%
		chatterjee2015matrix}
	\APACinsertmetastar {%
		chatterjee2015matrix}%
	\begin{APACrefauthors}%
		Chatterjee, S.%
	\end{APACrefauthors}%
	\unskip\
	\newblock
	\APACrefYearMonthDay{2015}{}{}.
	\newblock
	{\BBOQ}\APACrefatitle {Matrix estimation by universal singular value
		thresholding} {Matrix estimation by universal singular value
		thresholding}.{\BBCQ}
	\newblock
	\APACjournalVolNumPages{The Annals of Statistics}{43}{1}{177--214}.
	\PrintBackRefs{\CurrentBib}
	
	\bibitem [\protect \citeauthoryear {%
		Chen%
		, Fan%
		, Ma%
		\BCBL {}\ \BBA {} Yan%
	}{%
		Chen%
		\ \protect \BOthers {.}}{%
		{\protect \APACyear {2019}}%
	}]{%
		chen2019inference}
	\APACinsertmetastar {%
		chen2019inference}%
	\begin{APACrefauthors}%
		Chen, Y.%
		, Fan, J.%
		, Ma, C.%
		\BCBL {}\ \BBA {} Yan, Y.%
	\end{APACrefauthors}%
	\unskip\
	\newblock
	\APACrefYearMonthDay{2019}{}{}.
	\newblock
	{\BBOQ}\APACrefatitle {Inference and uncertainty quantification for noisy
		matrix completion} {Inference and uncertainty quantification for noisy matrix
		completion}.{\BBCQ}
	\newblock
	\APACjournalVolNumPages{Proceedings of the National Academy of
		Sciences}{116}{46}{22931--22937}.
	\PrintBackRefs{\CurrentBib}
	
	\bibitem [\protect \citeauthoryear {%
		Chi%
		, Zhou%
		, Chen%
		, Del~Vecchyo%
		\BCBL {}\ \BBA {} Lange%
	}{%
		Chi%
		\ \protect \BOthers {.}}{%
		{\protect \APACyear {2013}}%
	}]{%
		chi2013genotype}
	\APACinsertmetastar {%
		chi2013genotype}%
	\begin{APACrefauthors}%
		Chi, E\BPBI C.%
		, Zhou, H.%
		, Chen, G\BPBI K.%
		, Del~Vecchyo, D\BPBI O.%
		\BCBL {}\ \BBA {} Lange, K.%
	\end{APACrefauthors}%
	\unskip\
	\newblock
	\APACrefYearMonthDay{2013}{}{}.
	\newblock
	{\BBOQ}\APACrefatitle {Genotype imputation via matrix completion} {Genotype
		imputation via matrix completion}.{\BBCQ}
	\newblock
	\APACjournalVolNumPages{Genome research}{23}{3}{509--518}.
	\PrintBackRefs{\CurrentBib}
	
	\bibitem [\protect \citeauthoryear {%
		Cottet%
		\ \BBA {} Alquier%
	}{%
		Cottet%
		\ \BBA {} Alquier%
	}{%
		{\protect \APACyear {2018}}%
	}]{%
		cottet20181bit}
	\APACinsertmetastar {%
		cottet20181bit}%
	\begin{APACrefauthors}%
		Cottet, V.%
		\BCBT {}\ \BBA {} Alquier, P.%
	\end{APACrefauthors}%
	\unskip\
	\newblock
	\APACrefYearMonthDay{2018}{}{}.
	\newblock
	{\BBOQ}\APACrefatitle {1-Bit matrix completion: PAC-Bayesian analysis of a
		variational approximation} {1-bit matrix completion: Pac-bayesian analysis of
		a variational approximation}.{\BBCQ}
	\newblock
	\APACjournalVolNumPages{Machine Learning}{107}{3}{579--603}.
	\PrintBackRefs{\CurrentBib}
	
	\bibitem [\protect \citeauthoryear {%
		Davenport%
		, Plan%
		, Van Den~Berg%
		\BCBL {}\ \BBA {} Wootters%
	}{%
		Davenport%
		\ \protect \BOthers {.}}{%
		{\protect \APACyear {2014}}%
	}]{%
		Davenport14}
	\APACinsertmetastar {%
		Davenport14}%
	\begin{APACrefauthors}%
		Davenport, M\BPBI A.%
		, Plan, Y.%
		, Van Den~Berg, E.%
		\BCBL {}\ \BBA {} Wootters, M.%
	\end{APACrefauthors}%
	\unskip\
	\newblock
	\APACrefYearMonthDay{2014}{}{}.
	\newblock
	{\BBOQ}\APACrefatitle {1-bit matrix completion} {1-bit matrix
		completion}.{\BBCQ}
	\newblock
	\APACjournalVolNumPages{Information and Inference: A Journal of the
		IMA}{3}{3}{189--223}.
	\PrintBackRefs{\CurrentBib}
	
	\bibitem [\protect \citeauthoryear {%
		Devroye%
		, Gy{\"o}rfi%
		\BCBL {}\ \BBA {} Lugosi%
	}{%
		Devroye%
		\ \protect \BOthers {.}}{%
		{\protect \APACyear {2013}}%
	}]{%
		devroye2013probabilistic}
	\APACinsertmetastar {%
		devroye2013probabilistic}%
	\begin{APACrefauthors}%
		Devroye, L.%
		, Gy{\"o}rfi, L.%
		\BCBL {}\ \BBA {} Lugosi, G.%
	\end{APACrefauthors}%
	\unskip\
	\newblock
	\APACrefYear{2013}.
	\newblock
	\APACrefbtitle {A probabilistic theory of pattern recognition} {A probabilistic
		theory of pattern recognition}\ (\BVOL~31).
	\newblock
	\APACaddressPublisher{}{Springer Science \& Business Media}.
	\PrintBackRefs{\CurrentBib}
	
	\bibitem [\protect \citeauthoryear {%
		Giraud%
	}{%
		Giraud%
	}{%
		{\protect \APACyear {2021}}%
	}]{%
		giraud2021introduction}
	\APACinsertmetastar {%
		giraud2021introduction}%
	\begin{APACrefauthors}%
		Giraud, C.%
	\end{APACrefauthors}%
	\unskip\
	\newblock
	\APACrefYear{2021}.
	\newblock
	\APACrefbtitle {Introduction to high-dimensional statistics} {Introduction to
		high-dimensional statistics}.
	\newblock
	\APACaddressPublisher{}{Chapman and Hall/CRC}.
	\PrintBackRefs{\CurrentBib}
	
	\bibitem [\protect \citeauthoryear {%
		Gross%
	}{%
		Gross%
	}{%
		{\protect \APACyear {2011}}%
	}]{%
		gross2011recovering}
	\APACinsertmetastar {%
		gross2011recovering}%
	\begin{APACrefauthors}%
		Gross, D.%
	\end{APACrefauthors}%
	\unskip\
	\newblock
	\APACrefYearMonthDay{2011}{}{}.
	\newblock
	{\BBOQ}\APACrefatitle {Recovering low-rank matrices from few coefficients in
		any basis} {Recovering low-rank matrices from few coefficients in any
		basis}.{\BBCQ}
	\newblock
	\APACjournalVolNumPages{IEEE Transactions on Information
		Theory}{57}{3}{1548--1566}.
	\PrintBackRefs{\CurrentBib}
	
	\bibitem [\protect \citeauthoryear {%
		Han%
		\ \protect \BOthers {.}}{%
		Han%
		\ \protect \BOthers {.}}{%
		{\protect \APACyear {2014}}%
	}]{%
		han2014linear}
	\APACinsertmetastar {%
		han2014linear}%
	\begin{APACrefauthors}%
		Han, X.%
		, Wu, J.%
		, Wang, L.%
		, Chen, Y.%
		, Senhadji, L.%
		\BCBL {}\ \BBA {} Shu, H.%
	\end{APACrefauthors}%
	\unskip\
	\newblock
	\APACrefYearMonthDay{2014}{}{}.
	\newblock
	{\BBOQ}\APACrefatitle {Linear total variation approximate regularized nuclear
		norm optimization for matrix completion} {Linear total variation approximate
		regularized nuclear norm optimization for matrix completion}.{\BBCQ}
	\newblock
	\BIn{} \APACrefbtitle {Abstract and Applied Analysis} {Abstract and applied
		analysis}\ (\BVOL\ 2014).
	\PrintBackRefs{\CurrentBib}
	
	\bibitem [\protect \citeauthoryear {%
		Hastie%
		, Tibshirani%
		, Friedman%
		\BCBL {}\ \BBA {} Friedman%
	}{%
		Hastie%
		\ \protect \BOthers {.}}{%
		{\protect \APACyear {2009}}%
	}]{%
		hastie2009elements}
	\APACinsertmetastar {%
		hastie2009elements}%
	\begin{APACrefauthors}%
		Hastie, T.%
		, Tibshirani, R.%
		, Friedman, J\BPBI H.%
		\BCBL {}\ \BBA {} Friedman, J\BPBI H.%
	\end{APACrefauthors}%
	\unskip\
	\newblock
	\APACrefYear{2009}.
	\newblock
	\APACrefbtitle {The elements of statistical learning: data mining, inference,
		and prediction} {The elements of statistical learning: data mining,
		inference, and prediction}\ (\BVOL~2).
	\newblock
	\APACaddressPublisher{}{Springer}.
	\PrintBackRefs{\CurrentBib}
	
	\bibitem [\protect \citeauthoryear {%
		Herbster%
		, Pasteris%
		\BCBL {}\ \BBA {} Pontil%
	}{%
		Herbster%
		\ \protect \BOthers {.}}{%
		{\protect \APACyear {2016}}%
	}]{%
		herbster2016mistake}
	\APACinsertmetastar {%
		herbster2016mistake}%
	\begin{APACrefauthors}%
		Herbster, M.%
		, Pasteris, S.%
		\BCBL {}\ \BBA {} Pontil, M.%
	\end{APACrefauthors}%
	\unskip\
	\newblock
	\APACrefYearMonthDay{2016}{}{}.
	\newblock
	{\BBOQ}\APACrefatitle {Mistake bounds for binary matrix completion} {Mistake
		bounds for binary matrix completion}.{\BBCQ}
	\newblock
	\APACjournalVolNumPages{Advances in Neural Information Processing
		Systems}{29}{}{}.
	\PrintBackRefs{\CurrentBib}
	
	\bibitem [\protect \citeauthoryear {%
		Hsieh%
		, Natarajan%
		\BCBL {}\ \BBA {} Dhillon%
	}{%
		Hsieh%
		\ \protect \BOthers {.}}{%
		{\protect \APACyear {2015}}%
	}]{%
		hsieh2015pu}
	\APACinsertmetastar {%
		hsieh2015pu}%
	\begin{APACrefauthors}%
		Hsieh, C\BHBI J.%
		, Natarajan, N.%
		\BCBL {}\ \BBA {} Dhillon, I.%
	\end{APACrefauthors}%
	\unskip\
	\newblock
	\APACrefYearMonthDay{2015}{}{}.
	\newblock
	{\BBOQ}\APACrefatitle {PU learning for matrix completion} {Pu learning for
		matrix completion}.{\BBCQ}
	\newblock
	\BIn{} \APACrefbtitle {International conference on machine learning}
	{International conference on machine learning}\ (\BPGS\ 2445--2453).
	\PrintBackRefs{\CurrentBib}
	
	\bibitem [\protect \citeauthoryear {%
		Ji%
		, Liu%
		, Shen%
		\BCBL {}\ \BBA {} Xu%
	}{%
		Ji%
		\ \protect \BOthers {.}}{%
		{\protect \APACyear {2010}}%
	}]{%
		ji2010robust}
	\APACinsertmetastar {%
		ji2010robust}%
	\begin{APACrefauthors}%
		Ji, H.%
		, Liu, C.%
		, Shen, Z.%
		\BCBL {}\ \BBA {} Xu, Y.%
	\end{APACrefauthors}%
	\unskip\
	\newblock
	\APACrefYearMonthDay{2010}{}{}.
	\newblock
	{\BBOQ}\APACrefatitle {Robust video denoising using low rank matrix completion}
	{Robust video denoising using low rank matrix completion}.{\BBCQ}
	\newblock
	\BIn{} \APACrefbtitle {2010 IEEE computer society conference on computer vision
		and pattern recognition} {2010 ieee computer society conference on computer
		vision and pattern recognition}\ (\BPGS\ 1791--1798).
	\PrintBackRefs{\CurrentBib}
	
	\bibitem [\protect \citeauthoryear {%
		Jiang%
		\ \protect \BOthers {.}}{%
		Jiang%
		\ \protect \BOthers {.}}{%
		{\protect \APACyear {2016}}%
	}]{%
		jiang2016sparrec}
	\APACinsertmetastar {%
		jiang2016sparrec}%
	\begin{APACrefauthors}%
		Jiang, B.%
		, Ma, S.%
		, Causey, J.%
		, Qiao, L.%
		, Hardin, M\BPBI P.%
		, Bitts, I.%
		\BDBL {}Huang, X.%
	\end{APACrefauthors}%
	\unskip\
	\newblock
	\APACrefYearMonthDay{2016}{}{}.
	\newblock
	{\BBOQ}\APACrefatitle {SparRec: An effective matrix completion framework of
		missing data imputation for GWAS} {Sparrec: An effective matrix completion
		framework of missing data imputation for gwas}.{\BBCQ}
	\newblock
	\APACjournalVolNumPages{Scientific reports}{6}{1}{35534}.
	\PrintBackRefs{\CurrentBib}
	
	\bibitem [\protect \citeauthoryear {%
		Klopp%
		, Lafond%
		, Moulines%
		\BCBL {}\ \BBA {} Salmon%
	}{%
		Klopp%
		\ \protect \BOthers {.}}{%
		{\protect \APACyear {2015}}%
	}]{%
		klopp2015adaptive}
	\APACinsertmetastar {%
		klopp2015adaptive}%
	\begin{APACrefauthors}%
		Klopp, O.%
		, Lafond, J.%
		, Moulines, {\'E}.%
		\BCBL {}\ \BBA {} Salmon, J.%
	\end{APACrefauthors}%
	\unskip\
	\newblock
	\APACrefYearMonthDay{2015}{}{}.
	\newblock
	{\BBOQ}\APACrefatitle {Adaptive multinomial matrix completion} {Adaptive
		multinomial matrix completion}.{\BBCQ}
	\newblock
	\APACjournalVolNumPages{Electronic Journal of Statistics}{9}{}{2950--2975}.
	\PrintBackRefs{\CurrentBib}
	
	\bibitem [\protect \citeauthoryear {%
		Koren%
		, Bell%
		\BCBL {}\ \BBA {} Volinsky%
	}{%
		Koren%
		\ \protect \BOthers {.}}{%
		{\protect \APACyear {2009}}%
	}]{%
		koren2009matrix}
	\APACinsertmetastar {%
		koren2009matrix}%
	\begin{APACrefauthors}%
		Koren, Y.%
		, Bell, R.%
		\BCBL {}\ \BBA {} Volinsky, C.%
	\end{APACrefauthors}%
	\unskip\
	\newblock
	\APACrefYearMonthDay{2009}{}{}.
	\newblock
	{\BBOQ}\APACrefatitle {Matrix factorization techniques for recommender systems}
	{Matrix factorization techniques for recommender systems}.{\BBCQ}
	\newblock
	\APACjournalVolNumPages{Computer}{42}{8}{30--37}.
	\PrintBackRefs{\CurrentBib}
	
	\bibitem [\protect \citeauthoryear {%
		Lim%
		\ \BBA {} Teh%
	}{%
		Lim%
		\ \BBA {} Teh%
	}{%
		{\protect \APACyear {2007}}%
	}]{%
		lim2007variational}
	\APACinsertmetastar {%
		lim2007variational}%
	\begin{APACrefauthors}%
		Lim, Y\BPBI J.%
		\BCBT {}\ \BBA {} Teh, Y\BPBI W.%
	\end{APACrefauthors}%
	\unskip\
	\newblock
	\APACrefYearMonthDay{2007}{}{}.
	\newblock
	{\BBOQ}\APACrefatitle {Variational Bayesian approach to movie rating
		prediction} {Variational bayesian approach to movie rating
		prediction}.{\BBCQ}
	\newblock
	\APACjournalVolNumPages{Proceedings of KDD cup and workshop}{7}{}{15--21}.
	\PrintBackRefs{\CurrentBib}
	
	\bibitem [\protect \citeauthoryear {%
		Mai%
		\ \BBA {} Alquier%
	}{%
		Mai%
		\ \BBA {} Alquier%
	}{%
		{\protect \APACyear {2015}}%
	}]{%
		mai2015}
	\APACinsertmetastar {%
		mai2015}%
	\begin{APACrefauthors}%
		Mai, T\BPBI T.%
		\BCBT {}\ \BBA {} Alquier, P.%
	\end{APACrefauthors}%
	\unskip\
	\newblock
	\APACrefYearMonthDay{2015}{}{}.
	\newblock
	{\BBOQ}\APACrefatitle {A Bayesian approach for noisy matrix completion: Optimal
		rate under general sampling distribution} {A bayesian approach for noisy
		matrix completion: Optimal rate under general sampling distribution}.{\BBCQ}
	\newblock
	\APACjournalVolNumPages{Electron. J. Statist.}{9}{1}{823--841}.
	\newblock
	\begin{APACrefDOI} 10.1214/15-EJS1020 \end{APACrefDOI}
	\PrintBackRefs{\CurrentBib}
	
	\bibitem [\protect \citeauthoryear {%
		Mammen%
		\ \BBA {} Tsybakov%
	}{%
		Mammen%
		\ \BBA {} Tsybakov%
	}{%
		{\protect \APACyear {1999}}%
	}]{%
		mammen1999smooth}
	\APACinsertmetastar {%
		mammen1999smooth}%
	\begin{APACrefauthors}%
		Mammen, E.%
		\BCBT {}\ \BBA {} Tsybakov, A\BPBI B.%
	\end{APACrefauthors}%
	\unskip\
	\newblock
	\APACrefYearMonthDay{1999}{}{}.
	\newblock
	{\BBOQ}\APACrefatitle {Smooth discrimination analysis} {Smooth discrimination
		analysis}.{\BBCQ}
	\newblock
	\APACjournalVolNumPages{The Annals of Statistics}{27}{6}{1808--1829}.
	\PrintBackRefs{\CurrentBib}
	
	\bibitem [\protect \citeauthoryear {%
		Recht%
		\ \BBA {} R{\'e}%
	}{%
		Recht%
		\ \BBA {} R{\'e}%
	}{%
		{\protect \APACyear {2013}}%
	}]{%
		recht2013parallel}
	\APACinsertmetastar {%
		recht2013parallel}%
	\begin{APACrefauthors}%
		Recht, B.%
		\BCBT {}\ \BBA {} R{\'e}, C.%
	\end{APACrefauthors}%
	\unskip\
	\newblock
	\APACrefYearMonthDay{2013}{}{}.
	\newblock
	{\BBOQ}\APACrefatitle {Parallel stochastic gradient algorithms for large-scale
		matrix completion} {Parallel stochastic gradient algorithms for large-scale
		matrix completion}.{\BBCQ}
	\newblock
	\APACjournalVolNumPages{Mathematical Programming Computation}{5}{2}{201--226}.
	\PrintBackRefs{\CurrentBib}
	
	\bibitem [\protect \citeauthoryear {%
		Salakhutdinov%
		\ \BBA {} Mnih%
	}{%
		Salakhutdinov%
		\ \BBA {} Mnih%
	}{%
		{\protect \APACyear {2008}}%
	}]{%
		salakhutdinov2008bayesian}
	\APACinsertmetastar {%
		salakhutdinov2008bayesian}%
	\begin{APACrefauthors}%
		Salakhutdinov, R.%
		\BCBT {}\ \BBA {} Mnih, A.%
	\end{APACrefauthors}%
	\unskip\
	\newblock
	\APACrefYearMonthDay{2008}{}{}.
	\newblock
	{\BBOQ}\APACrefatitle {Bayesian probabilistic matrix factorization using Markov
		chain Monte Carlo} {Bayesian probabilistic matrix factorization using markov
		chain monte carlo}.{\BBCQ}
	\newblock
	\BIn{} \APACrefbtitle {Proceedings of the 25th international conference on
		Machine learning} {Proceedings of the 25th international conference on
		machine learning}\ (\BPGS\ 880--887).
	\PrintBackRefs{\CurrentBib}
	
	\bibitem [\protect \citeauthoryear {%
		Tsybakov%
	}{%
		Tsybakov%
	}{%
		{\protect \APACyear {2004}}%
	}]{%
		tsybakov2004optimal}
	\APACinsertmetastar {%
		tsybakov2004optimal}%
	\begin{APACrefauthors}%
		Tsybakov, A\BPBI B.%
	\end{APACrefauthors}%
	\unskip\
	\newblock
	\APACrefYearMonthDay{2004}{}{}.
	\newblock
	{\BBOQ}\APACrefatitle {Optimal aggregation of classifiers in statistical
		learning} {Optimal aggregation of classifiers in statistical
		learning}.{\BBCQ}
	\newblock
	\APACjournalVolNumPages{The Annals of Statistics}{32}{1}{135--166}.
	\PrintBackRefs{\CurrentBib}
	
	\bibitem [\protect \citeauthoryear {%
		Tsybakov%
		, Koltchinskii%
		\BCBL {}\ \BBA {} Lounici%
	}{%
		Tsybakov%
		\ \protect \BOthers {.}}{%
		{\protect \APACyear {2011}}%
	}]{%
		tsybakov2011nuclear}
	\APACinsertmetastar {%
		tsybakov2011nuclear}%
	\begin{APACrefauthors}%
		Tsybakov, A\BPBI B.%
		, Koltchinskii, V.%
		\BCBL {}\ \BBA {} Lounici, K.%
	\end{APACrefauthors}%
	\unskip\
	\newblock
	\APACrefYearMonthDay{2011}{}{}.
	\newblock
	{\BBOQ}\APACrefatitle {Nuclear-norm penalization and optimal rates for noisy
		low-rank matrix completion} {Nuclear-norm penalization and optimal rates for
		noisy low-rank matrix completion}.{\BBCQ}
	\newblock
	\APACjournalVolNumPages{Annals of Statistics}{39}{5}{2302--2329}.
	\PrintBackRefs{\CurrentBib}
	
	\bibitem [\protect \citeauthoryear {%
		Vapnik%
	}{%
		Vapnik%
	}{%
		{\protect \APACyear {1999}}%
	}]{%
		vapnik1999nature}
	\APACinsertmetastar {%
		vapnik1999nature}%
	\begin{APACrefauthors}%
		Vapnik, V.%
	\end{APACrefauthors}%
	\unskip\
	\newblock
	\APACrefYear{1999}.
	\newblock
	\APACrefbtitle {The nature of statistical learning theory} {The nature of
		statistical learning theory}.
	\newblock
	\APACaddressPublisher{}{Springer science \& business media}.
	\PrintBackRefs{\CurrentBib}
	
	\bibitem [\protect \citeauthoryear {%
		Zhang%
	}{%
		Zhang%
	}{%
		{\protect \APACyear {2004}}%
	}]{%
		zhang2004statistical}
	\APACinsertmetastar {%
		zhang2004statistical}%
	\begin{APACrefauthors}%
		Zhang, T.%
	\end{APACrefauthors}%
	\unskip\
	\newblock
	\APACrefYearMonthDay{2004}{}{}.
	\newblock
	{\BBOQ}\APACrefatitle {Statistical behavior and consistency of classification
		methods based on convex risk minimization} {Statistical behavior and
		consistency of classification methods based on convex risk
		minimization}.{\BBCQ}
	\newblock
	\APACjournalVolNumPages{The Annals of Statistics}{32}{1}{56--85}.
	\PrintBackRefs{\CurrentBib}
	
\end{thebibliography}
\end{document}